\pdfoutput=1
\documentclass[11pt]{article}

\usepackage[preprint]{coling}

\usepackage{times}
\usepackage{latexsym}

\usepackage[T1]{fontenc}
\usepackage[utf8]{inputenc}

\usepackage{microtype}

\usepackage{inconsolata}

\usepackage{graphicx}

\usepackage{float}
\usepackage{chngcntr}
\usepackage{amsmath}

\title{Personas with Attitudes: Controlling LLMs for Diverse Data Annotation}

\author{
 \textbf{Leon Fröhling\textsuperscript{1}},
 \textbf{Gianluca Demartini\textsuperscript{2}},
 \textbf{Dennis Assenmacher\textsuperscript{1}}\\
 \textsuperscript{1}GESIS - Leibniz Institute for the Social Sciences,
 \textsuperscript{2}The University of Queensland
\\
\texttt{\{leon.froehling, dennis.assenmacher\}@gesis.org, g.demartini@uq.edu.au}
}

\begin{document}
\maketitle
\begin{abstract}
We present a novel approach for enhancing diversity and control in data annotation tasks by personalizing large language models (LLMs). We investigate the impact of injecting diverse persona descriptions into LLM prompts across two studies, exploring whether personas increase annotation diversity and whether the impacts of individual personas on the resulting annotations are consistent and controllable. Our re  sults show that persona-prompted LLMs produce more diverse annotations than LLMs prompted without personas and that these effects are both controllable and repeatable, making our approach a suitable tool for improving data annotation in subjective NLP tasks like toxicity detection. 
% We presents a novel approach to enhancing diversity and control in data annotation tasks by personalizing large language models (LLMs) through the use of persona-based prompts. We investigate the impact of injecting diverse persona descriptions into LLM prompts across two studies, exploring whether personas increase annotation diversity and whether the impacts of individual personas on the resulting annotations are consistent and controllable. Our results show that persona-prompted LLMs produce more diverse annotations than the LLMs prompted without personas do, mimicking the subjective patterns found in human annotations. We further show that persona effects on annotations are both controllable and repeatable, making our approach a suitable tool for improving data annotation in subjective tasks like toxicity detection. Our findings demonstrate that the inclusion of persona descriptions in prompts for annotation tasks is successful in leading the LLMs to take a broad range of different (human) perspectives, offering a controlled method to improve data annotation and model training for NLP tasks requiring nuanced human judgment.
\end{abstract}

\textbf{Content Warning}: This document shows content that some may find disturbing, including content that is hateful towards protected groups.\\

\section{Introduction}

Many diverse NLP tasks rely on data acquired through human annotations, often collected via crowdworking studies. \citet{rottger2022two} recently offered a detailed introduction and discussion of the two different paradigms available to handle label differences resulting from diverse beliefs and backgrounds of individual annotators. The \textit{prescriptive} paradigm discourages annotator subjectivity and strives towards a single label for each instance in a dataset, effectively enabling the training of models that consistently apply a single understanding of the construct. The ease with which the resulting data may be used for model training makes this paradigm popular in NLP, often applying a simple majority vote to the annotations offered by a diverse pool of crowdworkers to derive a single ground-truth label for the annotated instance. Thereby, this paradigm is akin to the "wisdom of the crowd" approach, assuming that the "average opinion" of multiple, diverse annotators would sufficiently approximate the desired ground-truth label. 

The second and alternative paradigm discussed by \citet{rottger2022two} is the \textit{descriptive} paradigm, in which annotator subjectivity is explicitly encouraged and used to gain insights into diverse beliefs and to improve model training and evaluation via the consideration of annotator disagreement. \citet{rottger2022two} conclude that neither of the two paradigms is inherently superior, but that both serve different purposes and applications - for example, the prescriptive paradigm makes it easier to train classification models on the resulting data, while only the descriptive paradigm allows researchers to study and understand differences in perceptions across annotators with diverse beliefs and backgrounds.

In this work, we propose to combine the idea of using LLMs as annotators with the large pool of personas offered by the Persona Hub \citep{chan2024scaling} to both increase and control the diversity of the generated annotations. 
To explore the feasibility of injecting personas into LLM prompts to diversify and steer the models' zero-shot annotations in all its facets, we organize this work into two studies. Study 1 covers the \textit{prescriptive} paradigm towards diverse annotations, assuming the existence of a single label per instance and evaluating our approach's annotation diversity by comparing the persona-prompted LLM annotations to it. Study 2, in contrast, is in line with the \textit{descriptive} paradigm, exploring the approach's ability to reconstruct the diversity found in groups of human annotators and to controllably replicate the annotation differences found in groups of human annotators. While both studies individually test the suitability of our LLM persona-prompt approach for the corresponding paradigm, they - taken together - are designed to establish that our approach is suitable to increase the diversity in LLM annotations (Study 1) and that the persona impact is not entirely random, but that it follows predictable and controllable patterns (Study 2). In this paper, therefore, we set out to answer the following two research questions:

\begin{itemize}
    \item RQ Study 1: Does the inclusion of persona descriptions in LLM-prompts consistently increase the diversity of the resulting LLM annotations?
    \item RQ Study 2: Are there any patterns in the persona-prompted LLM annotations, and do these patterns align with effects of subjectivity observed for human annotators?
\end{itemize}

\section{Related Work}
% LLM annotations
Researchers have explored the abilities and performance of LLMs in annotating datasets for different types of constructs. \citet{ziems2024can}, \citet{faggioli2023perspectives}, \citet{gilardi2023chatgpt} and \citet{pavlovic2024effectiveness} provide overviews of scenarios for which the idea of using LLMs as annotators has already been tested, including tasks as diverse as determining the relevance of texts for specific issues, the detection of humor, or the extraction of medical information. Other researchers argued that LLMs are particularly well suited for the annotation of subjective constructs like hate speech, offensive language and toxicity \citep{li2024hot}. They argue that using LLMs can counter the instability in annotations that often arises from the varying social backgrounds of human annotators. Even more recently, researchers have started to explore the performance impacts of aggregating annotations generated with different LLMs \citep{del2024wisdom, schoenegger2024wisdom}, acting upon the assumption that an increased diversity of the crowd of annotators - be they human or LLM - would lead to gains in performance. \citet{he2024if} explore yet another angle of LLM annotations by comparing the annotation performance of GPT-4 with the annotations resulting from a carefully designed and conducted crowdworker annotation pipeline.  

% LLM annotations for subjective constructs
Yet another line of research is moving beyond the use of LLMs to predict the groundtruth label directly, proposing to \textit{personalize} LLMs via the inclusion of socio-demographic information in order to steer the LLM annotations towards those provided by human annotators. Among the most prominent approaches are \citet{argyle2023out}, \citet{bisbee2023synthetic} and \citet{santurkar2023whose}, who explore the ability of LLMs to predict survey responses of individual participants, as well as \citet{beck2024sensitivity}, \citet{pei2023annotator}, \citet{sun2023aligning} and \citet{orlikowski2023ecological}, who evaluate the performance of personalized LLMs in predicting the annotations of individual annotators as well as the resulting aggregated labels. 

% Synthetic data generation // Persona Hub
While the generation of dataset labels via LLMs might be interpreted as a form of synthetic training data generation, this description is usually reserved for efforts that synthetically create the instances to annotate, not (only) the corresponding labels. \citet{timpone2024artificial} offer an extensive review of the state of the literature together with a detailed discussion of associated challenges and opportunities. Fundamental for this work, \citet{chan2024scaling} introduce the Persona Hub, a collection of 1,000,000,000 diverse persona descriptions, as a way to diversify the synthetic instances that LLMs generate, and show that their persona descriptions - when included in the prompts used to synthesize, e.g., novel math or logical reasoning problems - are successful in increasing the diversity of the resulting datasets and thereby also the generalizability of the models trained on the datasets' tasks. 

\section{Data}
To systematically test the impact of including persona descriptions in the prompts used to collect toxicity annotations from different LLMs, we rely on two external sources of data described next.  

\subsection{Persona Descriptions}
Central to our proposal to increase the diversity of LLM-generated annotations via the injection of personas into the prompt is the collection of personas introduced by \citet{chan2024scaling} via their Persona Hub. While the personas themselves are just a brief, natural language description of an - ideally - human individual, the scale and diversity of the collection is what makes the Persona Hub such an ideal resource for our approach. \citet{chan2024scaling} developed the Persona Hub as part of a novel paradigm for the creation of synthetic data, not driven by seed datasets or manual prompt-design, but by a large number of personas to be automatically injected into LLM prompts. Their persona collection features brief descriptions of more than 1 billion different personas, created by asking different LLMs for a shown webtext instance: "who is likely to [read|write|like|dislike|...] this text". Depending on the prompt as well as the nature and level of detail of the webtext instance, the LLM will come up with different persona descriptions in response, varying both in content and complexity. Table~\ref{tab:personas} shows a selection of different personas included in the Persona Hub and the Appendix provides additional detail on the approach used by \citet{chan2024scaling} to create the persona descriptions as well as our efforts to clean them. 

\begin{table*}[ht]
    \footnotesize
    \centering
    \begin{tabular}{l|p{14cm}}
    ID & Persona \\
    \hline \hline
    189476 & An experienced biomedical engineer who has successfully brought cognitive rehabilitation devices to market \\
    11276 & A project manager who is skeptical about the practicality and cost-effectiveness of containerization \\
    123381 & A Muslim immigrant seeking legal assistance in defending their right to religious expression \\
    81722 & a French postgraduate student focusing on multidisciplinary research \\
    73592 & A retiree living in the Auckland suburbs who often has Sunday barbecues with her neighbors and enjoys discussing local issues with them. \\
    \end{tabular}
    \caption{Random sample of five persona descriptions in our dataset.}
    \label{tab:personas}
\end{table*}

\subsection{Toxicity Datasets}
As the basis for our experiments, we use the datasets prepared and shared by \citet{sap2022annotators}. To study the impact of human annotators' beliefs and identities on their annotations, \citet{sap2022annotators} curate datasets of potentially toxic texts with different characteristics. From the datasets shared by \citet{founta2018large} and \citet{vidgen2021learning}, they select those posts that are either vulgar, anti-Black, or detected to be African-American English. 
For their breadth of posts study, \citet{sap2022annotators} collect 571 posts that are annotated by 173 annotators (recruited from Amazon Mechanical Turk (MTurk)) for their levels of toxicity, amounting to six annotations per post. While annotators indicated the level of toxicity on a 5-point Likert scale, we first calculate the average level over the six annotations and then map this to a binary toxic/not toxic outcome, considering all instances with an average toxicity level larger than 2.5 as toxic. We use this dataset of 571 instances with binary toxicity labels resulting from a majority vote of six individual annotators as data basis for our Study 1 and refer to it as the \textsc{Single-Label Dataset}.
For their breadth of workers study, \citet{sap2022annotators} carefully select 12 posts that are then annotated for their levels of toxicity by a large population of 641 annotators. The annotators have been recruited from MTurk to be diverse along racial, political, and gender identities. Each annotator then annotated every post, (subjectively) answering the question "How toxic/hateful/disrespectful or offensive does this post seem to you?"~\footnote{In early experiments, we also tried the alternative question asked by \citet{sap2022annotators}, inquiring about the perceived toxicity of the post "to anyone" instead of "to you". Since early results did not differ much, we decided to focus on the "to you" variant of the question.}.  For our Study 2, we use the non-aggregated responses provided by the individual annotators, with responses for each instance ranging from "Not at all" (1) to "Very much so" (5) on a 5-point Likert scale. We refer to this dataset of 12 posts annotated by 641 annotators with labels from 1 to 5 as the \textsc{Multi-Label Dataset}.

\section{Methodology}
Experiments in all studies were conducted with two different LLMs from two popular families of open-source models - the Mistral-7B-Instruct-v0.1 model and the Qwen2-7B-Instruct model. Further justification for our choice of models and details on model deployment can be found in the Appendix.

\subsection{Study 1}
\label{sec:methodology_s1}
In Study 1, we want to explore RQ1 and establish that the inclusion of persona descriptions into the prompts used to generate LLM annotations increases the diversity of the models' annotation decisions, especially when compared to a baseline in which no persona description is added to the same prompt.

We randomly sample 1,000 personas from the 200,000 personas available in the Persona Hub \citep{chan2024scaling} for which we then collect their annotations on the \textsc{Single-Label Dataset}. We collect these persona-prompted LLM annotations by injecting the persona description directly into the prompt and asking for a binary label response using \textit{Prompt Template 1} shown in the Appendix. To compare the persona-prompted LLM annotations against LLM annotations without any persona-influence, we run the same models 1,000 times without any personas included in the prompt (also referred to throughout this paper as baseline LLM annotations) using \textit{Prompt Template 2} as shown in the Appendix. The variation in the no persona, baseline LLM setup is expected to originate exclusively from the randomness of the sampling process in generating the annotations across different runs of the same model. 

To further establish that the effect of including personas in LLM prompts are not randomly fluctuating, but that the inclusion of specific personas in prompts has a consistent effect on the resulting annotations, we select the 30 personas with the highest, median and lowest alignment to the \textsc{Single-Label Dataset} labels (as measured via the macro-average F1 score) and let them each annotate the dataset 30 additional times. 

To explore whether aggregating different persona-prompted LLM runs into crowds of annotators increases the alignment of the crowds' majority vote labels with the human majority vote labels, we randomly distribute each of the 1,000 personas and the 1,000 separate baseline runs into ten non-overlapping crowds of size 100. For these crowds, we track the performance trajectories resulting from a simple majority vote of the annotators in the crowd when systematically adding a single annotator to the pre-existing crowd at each step, resulting in artificial crowds ranging in size from one to 100 annotators. We acknowledge that the order in which the crowds are constructed has a potentially decisive impact on the crowds' trajectories, but argue that this is not unlike a practical scenario in which the crowd size would also be increased (e.g., by collecting the annotations from additional crowdworkers) without being able to directly control the crowd composition. To make explicit the impact of the order in which the crowds are constructed, we take a single set of 100 randomly sampled personas and randomly simulate 1,000 (of the theoretically available $100!$) different orders in which this one crowd can iteratively be built up.

\subsection{Study 2}
In Study 2, we turn towards the \textsc{Multi-Label Dataset} that holds a large number of different, subjective human perspectives expressed through different annotations for the same instances and construct.
To address RQ2 and explore the patterns that drive differences in annotations between different personas, we associate distances in the embedding space derived from persona descriptions with distances in the embedding space built upon the labelling decisions. For the creation of the persona embedding space, we use the 384-dimensional, pre-trained all-MiniLM-L12-v2~\footnote{https://huggingface.co/sentence-transformers/all-MiniLM-L12-v2} sentence-transformer model, trained to encode sentences and short paragraphs and thus well-suited to project our short persona descriptions into a vector space. The label embedding space is built directly upon the \textsc{Multi-Label Dataset} - each of the twelve instances in the dataset is represented as a dimension in the label embedding space, with values for each dimension ranging from 1 to 5, corresponding to the available label choice for each instance. To then derive the position of the personas in the label embedding space, we simply use each persona to annotate the \textsc{Multi-Label Dataset} following \textit{Prompt Template 3} shown in the Appendix, now soliciting toxicity levels on a 5-point Likert scale instead of as a binary label.

We start our analysis of the embedding spaces by using k-means to find clusters in the persona space, i.e., persona descriptions that are similar to each other. We then calculate the average distance in the label space between the personas in each of the resulting clusters with the personas of each other cluster, resulting in a symmetric matrix of average intra- and inter-cluster label space distances with dimensionality equal to the number of persona-space clusters. Based on that matrix, we can identify persona clusters that annotate alike as well as those that annotate very differently from each other. Additionally, we test the assumption that similar personas (i.e., small distance in the persona embedding space) annotate alike (i.e., small distance in the label embedding space) by calculating the pairwise persona distances in both spaces. For each persona, we calculate the pairwise distances to every other persona in both spaces, and then measure the correlation between these distances.

To test whether the annotation patterns we find for the persona-prompted LLM annotations are in line with the annotation patterns displayed by human annotators, we first formulate expectations of subjectivity effects based on the findings of \citet{sap2022annotators}. For their human MTurk annotators, they showed that conservative annotators were less likely to rate anti-Black posts as toxic but more likely to rate African-American English (AAE) posts as toxic and that Black annotators rate anti-Black posts as slightly more toxic than White annotators. While they derived from the literature that Black annotators would rate AAE posts as less toxic than White annotators, they could not find evidence for that. We propose to study these effects by comparing the annotations collected from personas that were not explicitly assigned to an ethnicity or an ideology via their descriptions to the annotations from versions of these neutral personas that we changed to be either Black or conservative via the injection of explicit markers (the terms "black" and "conservative") at manually selected, adequate positions in the persona descriptions. This results in three different groups, all based on the same set of neutral persona descriptions - ethnically and ideologically neutral personas (i.e., those originally identified for their neutrality), personas manually changed to be identifiable as Black and personas manually changed to be identifiable as conservative (see Appendix Table \ref{tab:s2_persona_groups} for example personas of each group). We then use these persona groups to annotate those subsets of the \textsc{Single-Label Dataset} that are identified as anti-Black or as AAE by \citet{sap2022annotators}, again soliciting annotations on the 5-point Likert scale introduced above.

\section{Results}
The following sections present the results of the experiments described above. All results can be reproduced using the code made available upon publication together with the publicly available datasets shared by \citet{sap2022annotators} and \citet{chan2024scaling}.~\footnote{The code to reproduce all results is available from https://github.com/frohleon/Personas-with-Attitudes.}

\subsection{Study 1}
\label{sec:results_s1}
\begin{figure}[t]
    \centering
    \includegraphics[width=\linewidth]{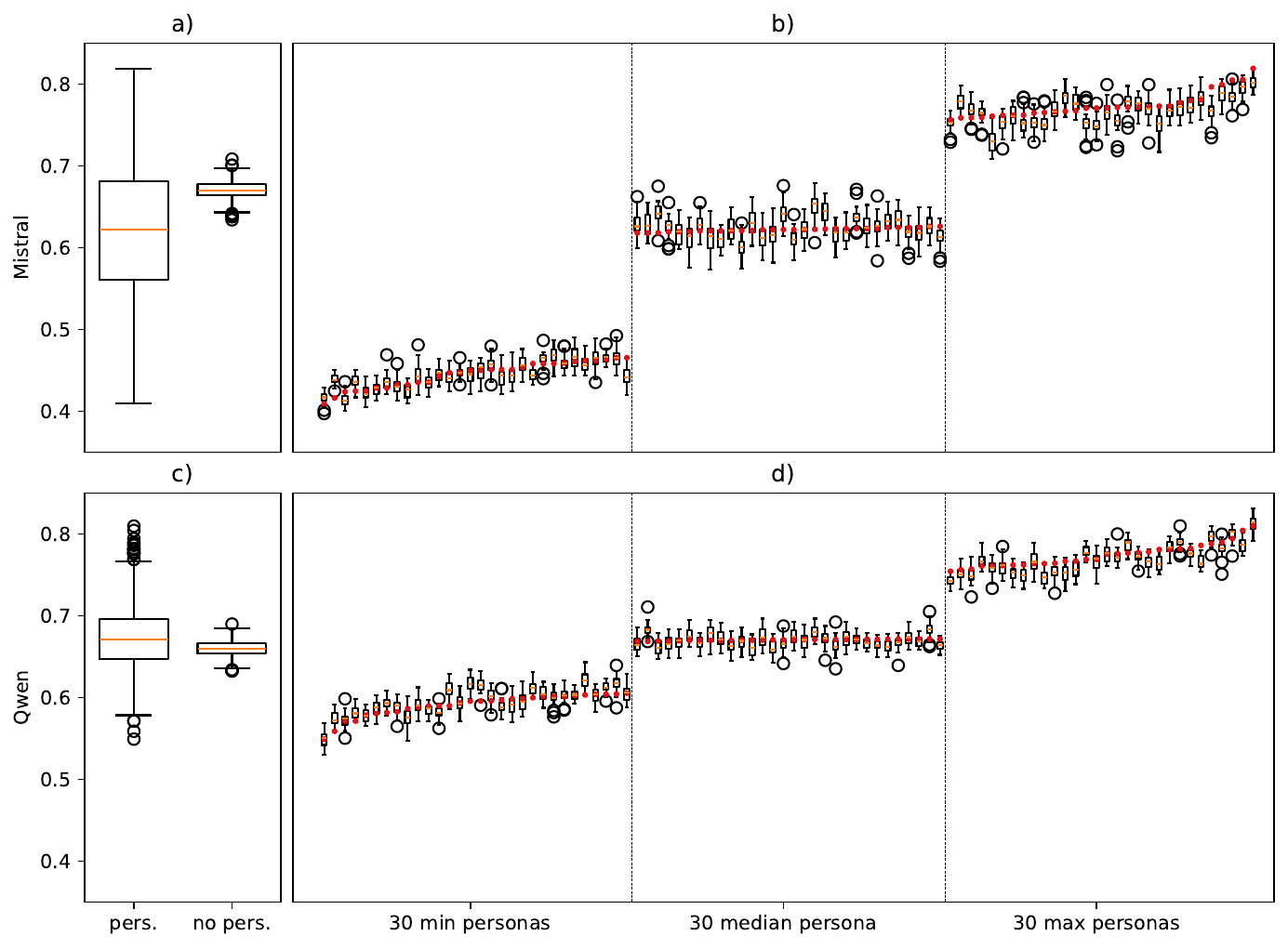}
    \caption{Boxplots of macro-average F1 scores achieved in 1,000 different persona-based LLM annotation (\textit{pers.}) and 1,000 baseline LLM annotation runs (\textit{no pers.}) for a) Mistral and c) Qwen. Boxplots of macro-average F1 scores achieved in 30 additional annotation runs for the 30 personas with min, median and max alignment to the human majority vote label for b) Mistral and d) Qwen.}
    \label{fig:s1_stability}
\end{figure}

When we examine the alignment of annotation runs with and without persona descriptions used in the prompt with the human majority vote labels represented by the \textsc{Single-Label Dataset}, the first thing to notice for both LLMs when including personas is the great increase in the diversity of alignment levels between the LLM and human annotators. Panels a) and c) in Figure \ref{fig:s1_stability} show boxplots of the distributions of the annotation performances (measured via macro-average F1 scores) resulting from the 1,000 persona-based and the 1,000 baseline annotation runs without personas. The boxplots indicate that for Mistral, the baseline LLM annotations are generally better aligned with the majority vote human annotations than the persona-prompted LLM annotations. On the contrary, LLM annotation runs with personas exhibit significantly more fluctuation in the resulting levels of alignment to the labels in the \textsc{Single-Label Dataset}, indicating a higher opinion diversity introduced by the persona descriptions. For Qwen, we observe that the median persona-prompted LLM annotation runs align slightly better with the human majority vote label than the median baseline LLM runs. Nonetheless and parallel to Mistral, we observe a much higher variance in annotation alignment for the persona-prompted than for the baseline runs.

This initial analysis of the various annotation runs leads to an important conclusion: the introduction of personas into LLM prompts broadens (or diversifies) the distribution of performances across annotation runs with both models, especially when compared to baseline runs.~\footnote{This finding is confirmed through a Levene test for equality of variances, which for both LLMs rejects the null hypothesis of equal variances at significance levels of $\alpha = 0.001$.} In other words, the personalization through persona descriptions shifts the LLMs further away from their baseline performance than the typical randomness introduced by the sampling procedure does.

After having established that the inclusion of personas in the model prompts is successful in widening the distribution of LLM annotations, we are next testing whether the effects of persona descriptions are consistent and stable across multiple runs and thus controllable, or whether the personas impact annotations randomly. Each boxplot in panels b) and d) of Figure \ref{fig:s1_stability} represents 30 annotation runs of the same persona, with individual runs differing only through the randomness of the sampling process in generating the annotations. For both models, we see how the order of the levels of alignment to the human majority vote label associated with different personas is almost perfectly restored when running each persona multiple times. We take this as a strong indication that the annotation differences associated with different personas are not purely contingent, but that the same personas consistently push the models into the same perspectives, impacting the annotations more strongly than the randomness of the sampling procedure in generating annotations. This is another important finding of the proposed persona-based annotation approach, as it establishes not just the consistency and stability of the persona-based annotation runs, but indicates also a degree of control that is required to steer the models towards specific annotation perspectives. 

In a first attempt to identify characteristics of personas that lead to particularly weak and strong alignment between persona-prompted LLM annotations and the human majority vote labels, we manually search for themes and patterns in the descriptions of the personas whose annotations are shown in Figure \ref{fig:s1_stability}. For Mistral, we find that personas with high alignment to the human majority vote labels are described as "appreciative", as "interested in" different questions and topics, as well as "offering" or "seeking advice". In contrast, for personas with low alignment, the term "competitive" occurs most frequently in the persona descriptions, together with expressions of "being against" something. Interestingly, the tendency that personas described as more open and outreaching achieve higher alignment than personas defined as fundamentally in opposition to something or someone is pretty much inverted for Qwen. There, we find that the personas described as "being critical", as "skeptical" or "questioning" of something have higher alignment, while the personas that "share" things or "seek" and "offer" advice have lower alignment. Tables \ref{tab:mistral_30} and \ref{tab:qwen_30} in the Appendix show the persona descriptions with the terms mentioned here highlighted. The opposing directions of the effect for the two models makes it inherently difficult to meaningfully interpret, but one certain conclusion is that character traits and psychological attributes seem to be more important for annotation diversity than socio-demographic attributes, at least for the extreme ends of the widened persona distributions.

\begin{figure}
    \centering
    \includegraphics[width=\linewidth]{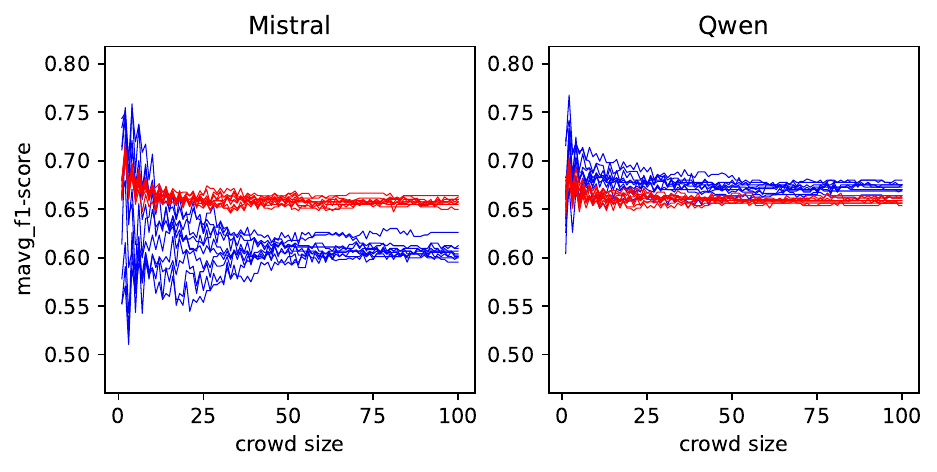}
    \caption{Macro-average F1 scores of majority vote performance for ten different persona-prompted LLM crowds (blue) and baseline LLM crowds (red) of sizes increasing from 1 to 100.}
    \label{fig:s1_crowds_mavgf1}
\end{figure}

When aggregating the annotations obtained from individual LLM annotation runs into majority vote annotations corresponding to LLM-crowds of different sizes, as shown in Figure \ref{fig:s1_crowds_mavgf1}, we do not observe closer alignment with increasing LLM-crowd size, as measured in terms of the macro-average F1 score achieved by the crowds' majority vote label against the human majority vote label represented in the \textsc{Single-Label Dataset}. Rather, we observe that the overall alignment for the crowds of persona-based LLMs fluctuates heavily with every annotator added, up to a crowd size of 50 (for Mistral) and 25 (for Qwen) before the resulting performances at each step stabilize. For the crowds made up from different baseline runs, i.e., without any personas, we do not observe such heavy fluctuations, but rather see a slow but steady deterioration of annotation alignment before stabilizing, starting with crowd sizes of between ten to 20 annotators. The fact that the crowd performances with personas fluctuate so much more strongly than those of baseline-run crowds speaks to the fact that the diversity in persona-based annotation runs is much higher than that in baseline runs. 
Simulating 1,000 permutations of the same crowds, i.e., changing only the order in which the crowds are constructed, confirms that this order has a larger effect for smaller crowd sizes, and again, much more so for persona-based crowds than baseline annotation crowds, but that the same patterns of convergence beyond crowd sizes of around 25 annotators result (see Appendix Figure \ref{fig:s1_crowds_permutations}).          

\subsection{Study 2}
\label{sec:results_s2}
\begin{figure}
    \centering
    \includegraphics[width=\linewidth]{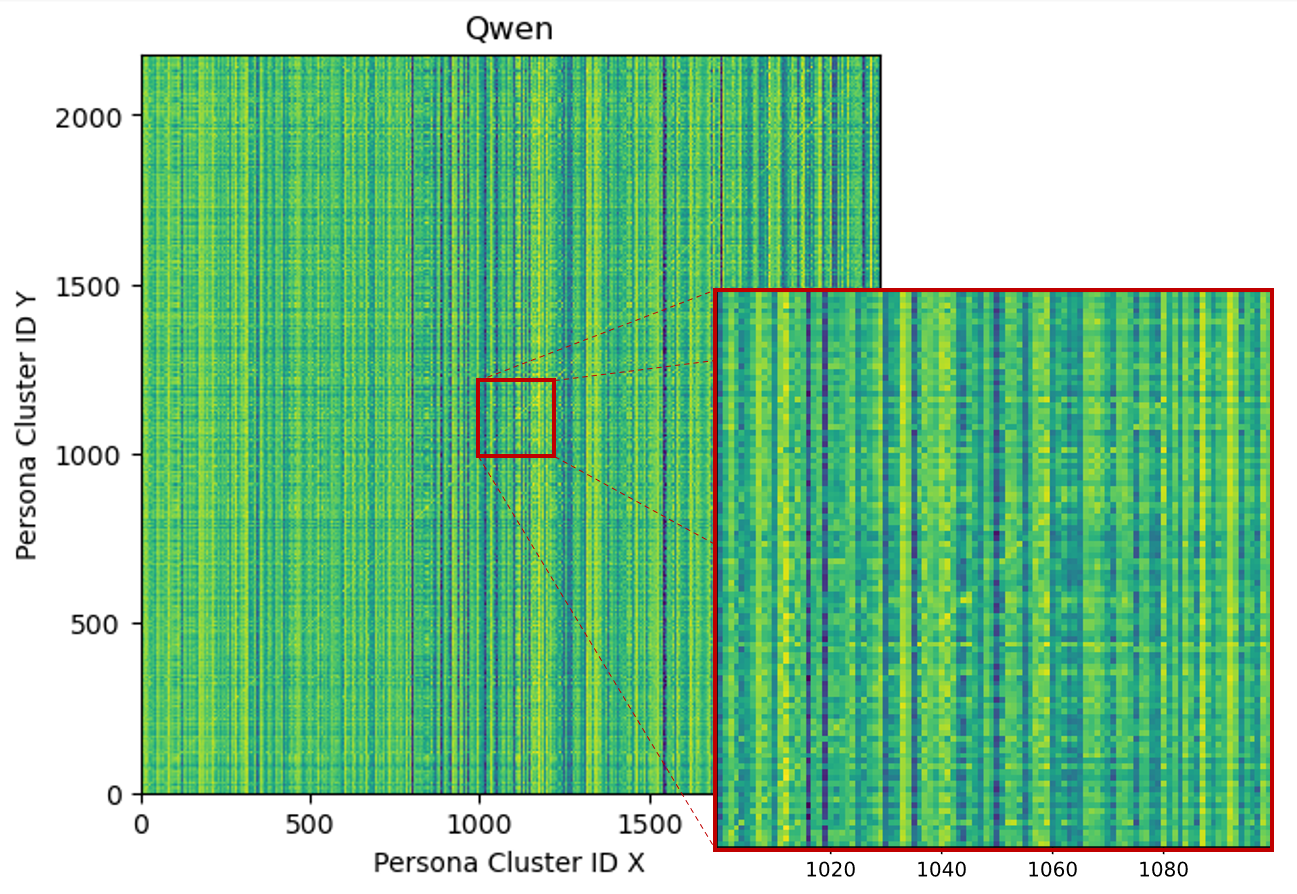}
    \caption{Intra- and inter-cluster cosine distances of persona-space clusters measured in label embedding space resulting from Qwen annotations. Values are normalized per row and lighter-colored cells represent lower average distances between the respective clusters. The inset zooms in on clusters with IDs from 1,000 to 1,100.}
    \label{fig:s2_pclusters_labelspace_qwen}
\end{figure}

After having established in Study 1 that different personas consistently lead to different levels of alignment with the human majority opinion, we are now taking a systematic look at the associations between persona descriptions and labelling decisions. We select a clustering solution in the persona embedding space with 2,180 different clusters covering 138,519 (70\% of all) personas, using a similarity threshold of 0.6 for cluster formation (see Appendix for justification and a basic evaluation of our clustering). Figure \ref{fig:s2_pclusters_labelspace_qwen} shows the intra- and inter-cluster cosine distances measured in the label embedding space resulting from Qwen annotations for the clusters found in the persona embedding space, and Appendix Figure \ref{fig:s2_pclusters_labelspace_mistral} shows the same for the label embedding space distances of Mistral annotations. For both models, but for Qwen a bit more pronounced than for Mistral, we see that the clusters along the diagonal are slightly lighter in color, indicating that personas that ended up in the same cluster based on their descriptions are also relatively close to each other in the label embedding space, i.e., personas with similar descriptions tend to annotate alike. Figure \ref{fig:s2_lcluster_personaspace} in the Appendix confirms this association by displaying persona description similarities across clusters found in the label embedding space. These first indications of a positive association between distances in the persona embedding space and the label embedding space are further confirmed by the pairwise correlation results shown in Appendix Figure \ref{fig:s2_hist_correlations}. For both models, more than 95\% of pairwise Spearman correlation coefficients between inter-persona distances measured in both spaces are significantly different from zero, with 75.5\% of these significant correlation coefficients being positive for Mistral. For Qwen, the number of significantly positive correlation coefficients for pairwise distances in the two spaces is with 88.3\% even higher. This is another central finding for our proposed approach, as it establishes that similar persona descriptions lead to similar annotation outcomes - yet another indication that the persona descriptions allow for control of the annotation perspectives taken by the model beyond purely random differences.     

\begin{table*}[t!]
    \footnotesize
    \centering
    \begin{tabular}{p{3cm}p{0.7cm}|p{9.9cm}p{0.7cm}}
        AAE Instance & Diff. & Anti-Black Instance & Diff. \\
        \hline \hline
        % whats up my \textit{n***} & -2.31 & I was raised in Idaho, amazing place. There is a reason they call it whiteaho. There was 1 single black in my town growing up. He was a halfie and wasn't too bad. Though still did the whole poor black me whine. I don't remember ever meeting a J3w my entire life until I left the state. But. The Mexicans did infiltrate and there are quite a few wetbacks there. Only downfall. & 1.09 \\
        whats up my \textit{n***} & -2.31 & I was raised in Idaho, amazing place. There is a reason they call it whiteaho. There was 1 single black in my town growing up. He was a halfie and wasn't too bad. Though still did the whole poor black me whine. [...] & 1.09 \\
        \hline
        ma brotha ma \textit{n***} & -1.36 & I always hated visiting \textit{n***} York, but if I'm ever forced to I'll be sure to pack all my Confederate clothes and flags and to dance around Central Park! ( I may need a little protection) & 0.69 \\
        \hline
        % I don't even wanna get married but you know if my \textit{n***} propose to me imma say yes plz & -1.16 & Side note, can anyone explain why racism is necessarily bad? Humanity has survived for a few thousand years without the civil rights movement and we've done perfectly fine. Only on the last 60 years has it been an issue for the SOME PEOPLE. & 0.53 \\
        I don't even wanna get married but you know if my \textit{n***} propose to me imma say yes plz & -1.16 & Side note, can anyone explain why racism is necessarily bad? [...] Only on the last 60 years has it been an issue for the SOME PEOPLE. & 0.53 \\
        \hline
    \end{tabular}
    \caption{AAE and anti-Black instances with largest difference in average toxicity ratings assigned by Black and conservative personas. Differences calculated as $\text{Diff}=\mu_{Q;B}-\mu_{Q;C}$ based on Qwen annotations. We replace all mentions of the n-word and its variants with \textit{n***} and mark omissions due to length constraints with [...].}
    \label{tab:s2_q_diff_examples}
\end{table*}

\begin{figure}
    \centering
    \includegraphics[width=\linewidth]{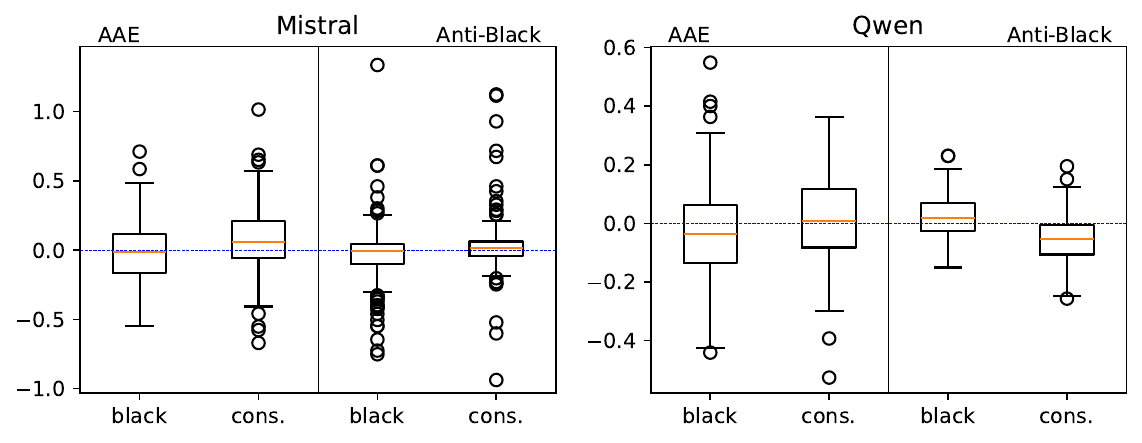}
    \caption{Boxplots of shifts in average toxicity labels assigned to instances in the AAE and anti-Black datasets. The shifts are on a persona-level and are calculated as the difference in average toxicity label of the manually changed black and conservative personas relative to the original, neutral persona.}
    \label{fig:s2_diff_boxplots}
\end{figure}

To test whether the impact of changing the originally (racially and ideologically) neutral persona descriptions to be clearly marked as either Black or conservative is in line with theoretical expectations, Figure \ref{fig:s2_diff_boxplots} shows that the effect on the annotations is relatively small, with mean toxicity level shifts across all instances in the AAE and anti-Black datasets and across all personas of the two groups close to zero.\footnote{Wilcoxon rank-sum tests for the theorized effects confirm that only the effects for Qwen on the anti-Black instances are statistically significant at significance levels of $\alpha = 0.05$.} However, for the AAE instances we observe for both models a slight shift in toxicity levels that is in line with theory, i.e., that personas marked as Black tend to perceive these instances as less toxic, both when compared to the neutral original personas as well as in comparison to the shifts observed for the conservative personas relative to the neutral baseline. For Qwen, the mean (absolute) toxicity level assigned to AAE instances across Black personas is $\mu_{Q;B}=3.39$ and $\mu_{Q;C}=3.43$ across conservative personas. For Mistral, the values are $\mu_{M;B}=2.67$ and $\mu_{M;C}=2.76$. For the anti-Black instances, we would expect Black personas to rate them as more toxic, a direction that is to a small degree confirmed by the distribution of differences in average toxicity levels of Black personas and their neutral counterparts for Qwen. While for Qwen, Black personas tend to annotate anti-Black posts as slightly more toxic than their neutral counterparts, conservative personas tend to annotate the same posts as less toxic than the original, neutral personas. For Mistral, the shifts caused by the inclusion of markers for Blackness and conservatism are far less pronounced on the anti-Black posts and are, if different from zero at all, shifted in a direction that is not in line with the theorized connection, i.e., such that Black personas perceive anti-Black posts as more toxic and conservative personas perceive them as less toxic. For Qwen, the mean (absolute) toxicity level assigned to anti-Black instances across Black personas is $\mu_{Q;B}=4.69$ and $\mu_{Q;C}=4.61$ across conservative personas. For Mistral, the values are $\mu_{M;B}=4.51$ and $\mu_{M;C}=4.58$.
While the quantitative shifts are relatively small and not in all settings entirely conclusive, a closer look at some of the instances for which Black and conservative annotators differ in their toxicity annotations most strongly is instructive. Table \ref{tab:s2_q_diff_examples} shows the three instances in the AAE dataset for which the differences between the absolute average toxicity level assigned by conservative and Black personas via the Qwen model are the largest. All of the shown instances for which Black annotators assigned a (much) lower toxicity level than their conservative counterparts are examples of a reclaimed usage of the n-word and thus examples of an explicitly non-toxic usage of a term usually used as a slur - a finding that perfectly mirrors what \citet{sap2022annotators} observe for human annotators, where "raters who are more conservative tend to score those posts [containing the n-word] as significantly more racist". This indicates that the inclusion of the Black marker in the persona prompt triggers an awareness in the model for the possible use of the n-word in a reclaimed, colloquial manner, a usage that should not be annotated as toxic. Table \ref{tab:s2_q_diff_examples} further shows the three instances in the anti-Black dataset for which the differences between the absolute toxicity level assigned by Black and conservative personas were the largest - i.e., those instances, for which Black personas on average assigned a higher absolute toxicity level than conservative personas did. In contrast to the AAE instances that have been annotated as less toxic by Black personas, these instances here are blatantly racist, thus serving well to explain why personas marked as Black would perceive them as particularly toxic, especially when compared to the ratings assigned by the conservative personas. Both observations are also true for annotations generated via Mistral, as shown in Appendix Table \ref{tab:s2_m_diff_examples}.

\section{Discussion and Conclusion}
In this work we explored the potential of personalizing LLMs through persona-based prompts to enhance diversity and control in data annotation tasks. By injecting persona descriptions into LLM prompts, we observed an increase in the variability of model annotations compared to annotation runs that did not include persona descriptions, demonstrating through various experiments that personas can influence model outputs in a consistent and controllable manner. We show that our persona-based approach to LLM data annotations offers a novel way to simulate human subjectivity in annotations, which can be particularly useful in tasks that require diverse and subjective perspectives, such as the detection of toxicity.

Our findings suggest that personas not only introduce desirable diversity in annotations, but that they also enable researchers to guide LLMs toward specific annotation behaviors, making them - under certain conditions - more aligned with groups of human annotators and being successful in replicating effects of annotation subjectivity also found in human annotations.

%\clearpage
\section{Limitations}
Our study is not without limitations. First, we restricted our analysis to two open-source LLMs. While we intended to include other models, such as LLaMA 3.1 and Falcon, different challenges unconnected to our proposed approach made their use for our purpose impossible - for various LLaMA models, the guardrails stopped the model from consistently complying with the toxicity annotation task, and for Falcon, the model's general ability to comply with the prompt instructions was insufficient for producing meaningful annotations. While our study establishes that the injection of personas into LLM prompt leads to the same effect of widening the annotation performance distribution across different models, future research could still investigate less complex subjective constructs, constructs that do not depend on potentially harmful language (e.g., sentiment detection), as well as additional model families, including those with (strong) guardrails.

Additionally, there are several limitations that we inherit from our use of the Persona Hub \citep{chan2024scaling} dataset. Importantly, our experimental study was conducted on a sample provided by the authors rather than the full dataset. This limitation may introduce sampling biases, potentially capturing certain demographic groups in the sample while excluding others, thereby potentially reducing the diversity effects observable in our analyses to the diversity captured in the persona sample. Furthermore, we cannot guarantee that all persona descriptions included in the sample represent individual humans (rather than groups of individuals or non-human characters like animals or even objects) and are written in meaningful English, although we took measures to filter out any persona descriptions written in languages other than English. Importantly, we do not have any control over the focus and make up of the persona descriptions. While this is not a necessary condition for our goal of showing that persona descriptions increase annotation diversity, we speculate that control over the information included in the descriptions would probably even lead to more significant effects than what we observed. We note that many of the personas have professions or hobbies as the most important descriptor, which are probably less important dimensions along which perceptions of toxicity differ than, e.g., dimensions such as race, gender or political ideology. Future research could explore the annotation effects caused by personas that differ along dimensions that are known from theory to be important factors for the annotation task at hand.

\bibliography{custom}

\appendix
\counterwithin{figure}{section}
\counterwithin{table}{section}

\section{Appendix}
\label{sec:appendix}
This Appendix is organized in sections that provide additional details on the Persona Hub dataset we use, the LLM prompts designed to collect (personalized) annotations from the models, the model deployment, as well as the results of the two experimental studies.

\subsection{Personas}
\label{app:personas}

\begin{table*}
    \footnotesize
    \centering
    \begin{tabular}{p{1.5cm}|p{12cm}}
         Persona ID & Neutral Persona with Replacement Token \\
         \hline \hline
         130831 & [ATOKEN] political science professor writing their first book about democracy \\
         164597 & [ATOKEN] receptionist at a boutique hotel who hates fake news \\
         82521 & An internationally recognized [TOKEN] car restoration expert with a web-based reality show \\
         \hline
    \end{tabular}
    \caption{Persona descriptions selected for their undefined ethnicity and ideology. These descriptions are changed into Black and conservative personas by replacing [ATOKEN] with "a black" and "a conservative" and [TOKEN] with "black" and "conservative", respectively.}
    \label{tab:s2_persona_groups}
\end{table*}

\citet{chan2024scaling} use two different approaches to automatically create personas from webtext (i.e., large-scale collections of text supposed to represent \textit{all text on the web}); text-to-persona, as described in the main part above, as well as persona-to-persona, an approach designed to complete the persona collection by leading the persona-generating LLM to consider personas beyond those visible and represented in the web, e.g., children, via their relations to the personas obtained from the text-to-persona approach. Their persona-to-persona prompt asks for any already created persona "who is in close relationship with the given persona" for up to six iterations, thereby enriching and diversifying the initial persona collection. Personas are then de-duplicated based on embedding proximity as well as ngram-overlaps.

In our experiments on crowd size and annotation diversity, we use the 200,000 personas that are publicly available (as of 03.07.2024). However, we noticed that some of these personas were written in languages other than English. Since we design our experiments assuming that the persona to be injected into the prompt is in English and since we do not (explicitly) leverage a multilingual LLM for the annotations, we make an effort to filter out any persona descriptions that are not written in English. To do so, we use the langdetect package~\footnote{https://pypi.org/project/langdetect/} to get a list of persona descriptions that the classifier labels as primarily written in a language other than English. Since we noticed that the classifier showed a slight tendency of producing false positives, i.e., flagging descriptions written in English as non-English, we manually went through this list and only selected those for removal that were clearly written in a language other than English. This cleaning process left us with a dataset of 198,039 different personas. During our experiments with the persona collection, we also noticed that some of the descriptions did not describe individual human personas, but groups of individuals or institutions. In one case, we even came across a persona description corresponding to a Brazilian spider. However, given the extremely low prevalence of these non-personas, the difficulty of finding and removing them, as well as the assumption that groups of individuals or institutions diversify annotations in a similar manner and direction as the single individuals they represent would do, we accept and acknowledge the inclusion of some non-personas as a limitation of our work.

For Study 2, we select and modify neutral personas to isolate and measure the effects associated with different annotator characteristics. Table \ref{tab:s2_persona_groups} shows three example personas manually selected for being ethnically and ideologically undefined, or neutral, as well as versions of the same personas that are explicitly marked as Black and conservative.

\subsection{LLM Prompts}
\label{app:llm_prompts}
This section shows the prompts used for soliciting the toxicity annotations from the LLMs. In each prompt, the [text]-token marks where the text to be annotated is inserted into the prompt. In prompts that are persona-injected, the [persona]-token marks where the persona description is inserted. 
In Study 1, we were interested in binary toxicity labels, using the following prompts for annotations with (Prompt Template 1) and without (Prompt Template 2) personas:

\begin{quote}
    \textit{Prompt Template 1}\\
    <TASK> Your task is to annotate the post shown below for toxicity, taking on the perspective of the following persona: [persona]. Respond with 'TRUE' if this post seems toxic/hateful/disrespectful to you and 'FALSE' if this post does not seem toxic/hateful/disrespectful to you. <POST> [text]. <RESPONSE> 
\end{quote}

\begin{quote}
\textit{Prompt Template 2}\\
    <TASK> Your task is to annotate the post shown below for toxicity. Respond with 'TRUE' if this post seems toxic/hateful/disrespectful to you and 'FALSE' if this post does not seem toxic/hateful/disrespectful to you. <POST> [text]. <RESPONSE> 
\end{quote}

In Study 3, we were interested in levels of toxicity on a 5-point Likert scale, using the following prompts for annotations with personas:

\begin{quote}
    \textit{Prompt Template 3}\\
    <TASK> Your task is to annotate the post shown below, taking on the perspective of the following persona: [persona]. Answer the following question on a scale from 1 to 5, where 1 stands for 'Not at all' and 5 stands for 'Very much so'. How toxic/hateful/disrespectful or offensive does this post seem to you? <POST> [text]. <RESPONSE> 
\end{quote}

\subsection{Model Deployment}
We used the Mistral-7B-Instruct-v0.1 model~\footnote{https://huggingface.co/mistralai/Mistral-7B-v0.1} and the Qwen2-7B-Instruct model~\footnote{https://huggingface.co/Qwen/Qwen2-7B-Instruct} via the Python bindings~\footnote{https://github.com/abetlen/llama-cpp-python} of the \texttt{llama.cpp} framework~\footnote{https://github.com/ggerganov/llama.cpp}, running them on two separate 40GB partitions of a NVIDIA A100 GPU. We choose to run all experiments on two separate models to establish that our approach is not dependent on a specific model, and selected the two models for their good zero-shot performance in the toxicity annotation task as well as their comparably good efficiency and usability, fitting on a single 40GB GPU partition.To further decrease the compute workload and streamline model generations, we make use of the Outlines framework~\footnote{https://github.com/outlines-dev/outlines}, effectively restricting the LLM generations to a provided set of response options. Since we are not interested in the LLMs' abilities to generate open ended responses but in their preferred alternative from a restricted set of options (either binary toxicity labels or the five ordinal options from the 5-point Likert scale for the level of toxicity), we consider this to be a sensible choice that does not impact the validity of our results. We use the multinomial sampler implemented in Outlines with a temperature of 1 for all generations.

\subsection{Study 1 Results}
\label{app:study1_results}
In this section, we report on the effect that the random permutation of the crowd composition order has on the annotation trajectories, and show additional performance metrics measured for the various experiments in Study 1.

\begin{figure}
    \centering
    \includegraphics[width=\linewidth]{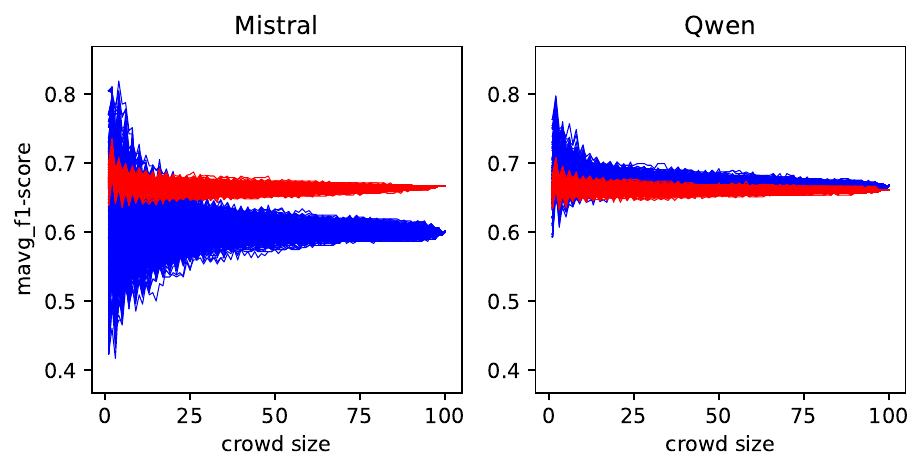}
    \caption{Macro-average F1 score of majority vote performance for 1,000 permutations of the same persona-prompted LLM crowd (blue) and baseline LLM crowd (red) of sizes increasing from 1 to 100.}
    \label{fig:s1_crowds_permutations}
\end{figure}

Figure \ref{fig:s1_crowds_permutations} reveals that the order of crowd build-up has an impact on the annotation alignment for LLM-crowds of smaller sizes, but that the levels of annotation alignment with the human majority vote labels converges for all crowd compositions from crowd sizes of 15 to 20 for Qwen and around 25 for Mistral.

\begin{figure*}
    \centering
    \includegraphics[width=\linewidth]{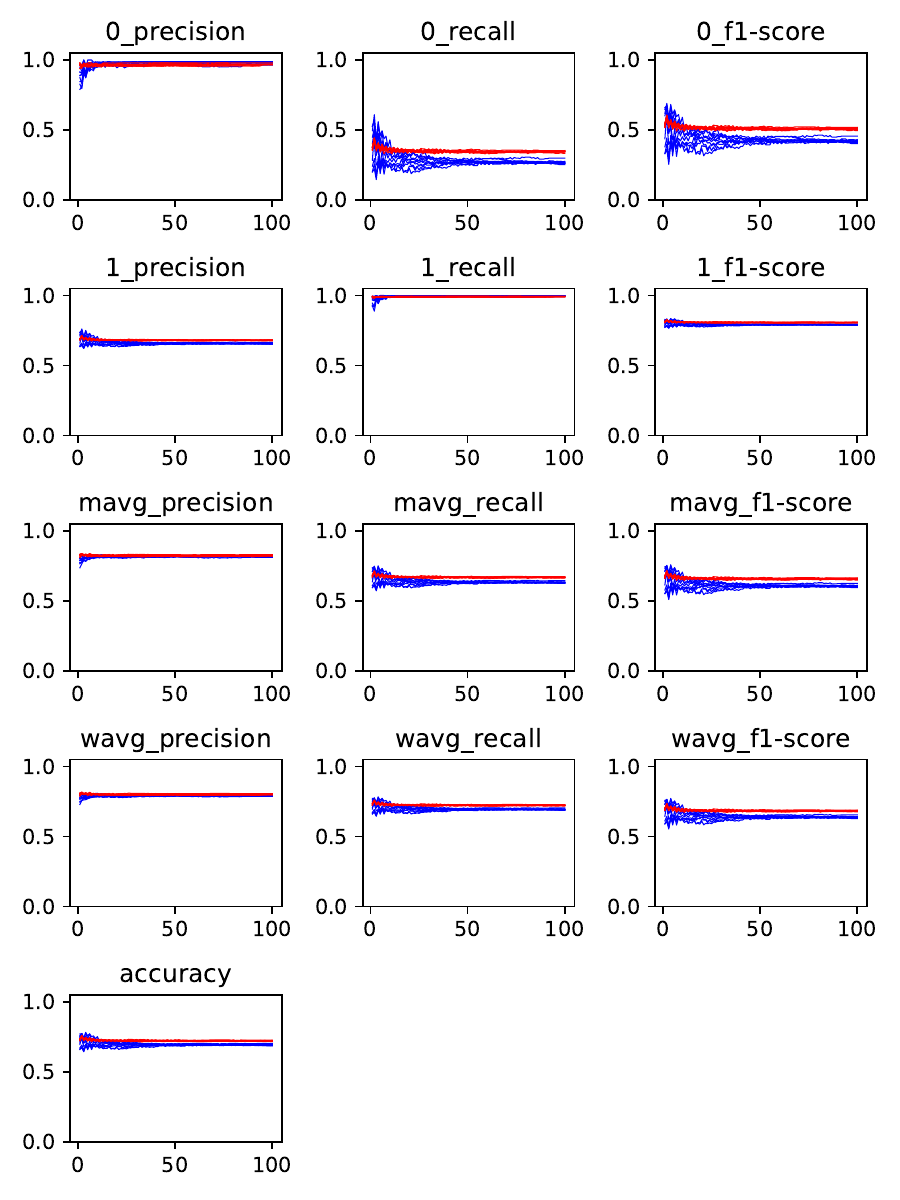}
    \caption{Different performance metrics for annotations collected from Mistral in ten different persona-prompted LLM crowds (blue) and baseline LLM crowds (red) of sizes increasing from 1 to 100. Here and for in following figures, \textit{0} refers to the \textit{no toxicity} labels and \textit{1} refers to the \textit{toxicity} labels, \textit{mavg} stands for \textit{macro-average} and \textit{wavg} for \textit{weighted-average}.}
    \label{fig:m_app_s1_crowds}
\end{figure*}

\begin{figure*}
    \centering
    \includegraphics[width=\linewidth]{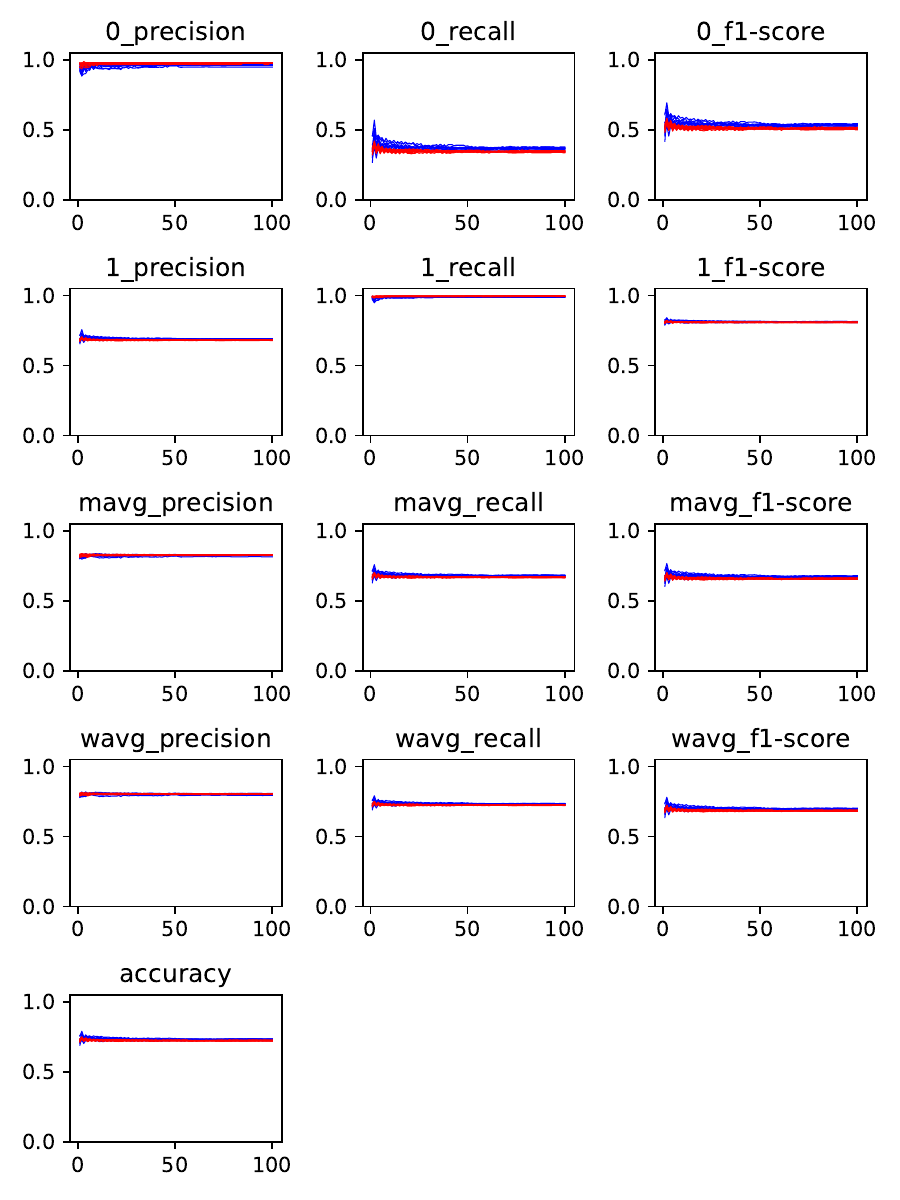}
    \caption{Different performance metrics for annotations collected from Qwen in ten different persona-prompted LLM crowds (blue) and baseline LLM crowds (red) of sizes increasing from 1 to 100.}
    \label{fig:q_app_s1_crowds}
\end{figure*}

\begin{figure*}
    \centering
    \includegraphics[width=\linewidth]{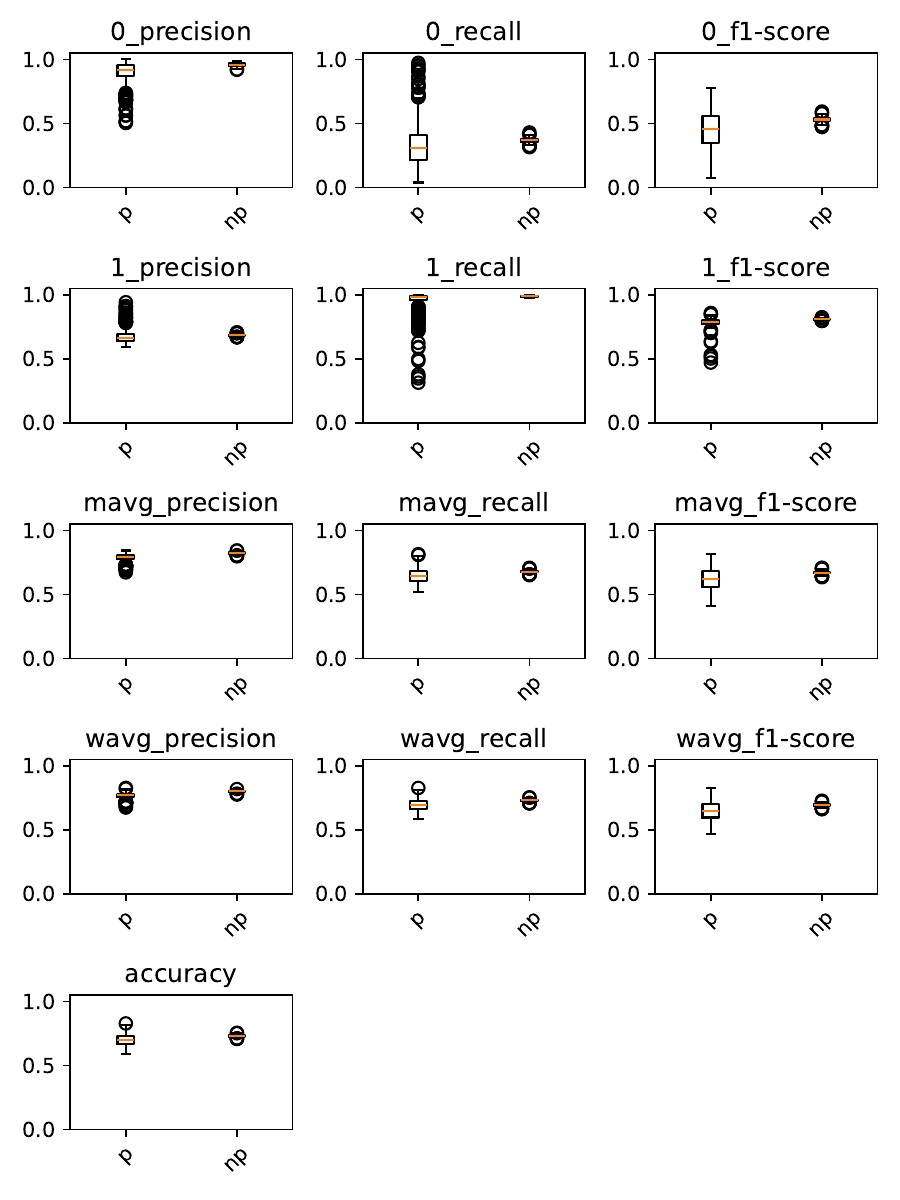}
    \caption{Boxplots of the distribution of different performance metrics resulting from 1,000 individual persona-prompted LLM annotation runs (p) and baseline LLM annotation runs (np) with Mistral.}
    \label{fig:m_app_s2_boxplots}
\end{figure*}

\begin{figure*}
    \centering
    \includegraphics[width=\linewidth]{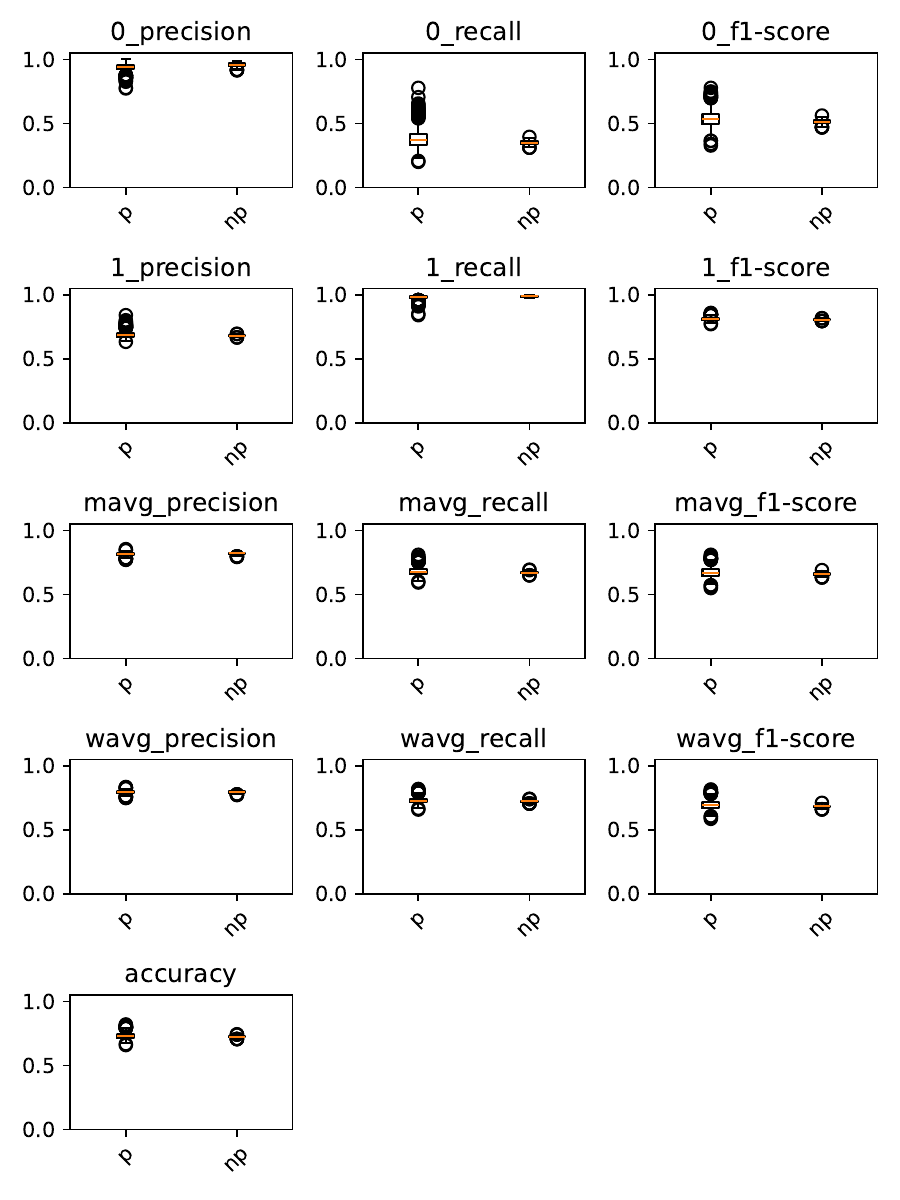}
    \caption{Boxplots of the distribution of different performance metrics resulting from 1,000 individual persona-prompted LLM annotation runs (p) and baseline LLM annotation runs (np) with Qwen.}
    \label{fig:q_app_s2_boxplots}
\end{figure*}

Figures \ref{fig:s1_crowds_permutations} to \ref{fig:q_app_s2_boxplots} show further performance metrics measured for the alignment between the majority vote labels from crowds of persona-prompted LLMs and baseline LLMs as well as individual runs from persona-prompted and baseline LLMs with the human majority vote labels as manifested in the \textsc{Single-Label Dataset}.

\begin{table*}
    \footnotesize
    \centering
    \begin{tabular}{p{6.5cm}|p{9cm}}
        relieved math teacher who is \underline{startled by} the received request & A photographer \underline{captivated by} the researcher's creative process and \underline{inspired to} capture their watercolor art in stunning photographs\\
        A \underline{tensioned}, deadline-driven managing editor who relies heavily on her team to prevent any mistakes & A parent who is concerned about their child's digestive issues and \underline{seeks advice} on probiotics and gut-friendly foods\\
        A photographer for a \underline{rival} camera brand constantly trying to outdo the representative's equipment & A physician who incorporates medical cannabis as part of their treatment plan for certain patients\\
        A communications director for a \underline{competing} football club aiming to dominate the regional sports coverage & A resilient woman who mobilizes her community to provide shelter and resources for displaced families\\
        A retired English teacher who \underline{disapproves} modern news broadcast & A fellow gallery owner who focuses on abstract expressionism and \underline{appreciates} the innovation of the robot sketches\\
        a person who is \underline{not a fan of} rich people & A behavioral therapist specializing in autism spectrum disorders, \underline{guiding} the parent in implementing effective strategies at home\\
        A \underline{disgruntled} dealer who was fired for suspected collusion with players & A digital storyteller who creates animated videos exploring the cultural significance of folklore from around the world\\
        A \underline{competing} speechwriter known for their sharp and witty remarks & A laissez-faire community moderator who \underline{values} open dialogue and minimum intervention\\
        a person who has \underline{never won anything} in their life and generally believes lotteries are a waste of money & A modern techno DJ \underline{appreciating} different music genres and open to \underline{appreciate} their partner's taste\\
        a Portuguese grave caretaker who gets easily confused with unclear messages & A talented costume designer who \underline{provides helpful tips and advice} on creating the perfect Rocket Raccoon outfit\\
        A resident of a small coastal town who believes the treasure hunting \underline{disrupts} the community & I am an elderly librarian with a \underline{passion for} onomatology (the study of the history and origin of names) and an \underline{affinity for} asteroids and astrology.\\
        A resourceful scavenger who believes Carl is taking up valuable supplies & A local resident who leads historical tours in the region, sharing anecdotes and legends about the military officer's ancestors\\
        A fellow sports mentor with a completely different training philosophy & A college student with limited knowledge about lymphedema but a lot of \underline{interest in} finding unexpected applications or uses for usual things.\\
         A tech-savvy supervillain who \underline{constantly challenges} the apprentice's skills and knowledge & I am a historical fiction novelist, deeply \underline{fascinated by} the complex and often untold narratives of women who held power in eras that were predominantly male-dominated.\\
        A prominent politician who \underline{dislikes being scrutinized} by the media & A geneticist who can \underline{provide information} about the latest research and treatment options for the hereditary disease\\
        A \underline{diligent and ambitious} sergeant who often reflects on shared past cases for learning & A large corporation \underline{seeking} a comprehensive business automation solution to optimize their various departments\\
        a resident of Yalım who has lived there for 60 years. & A renowned film critic who \underline{appreciates} the unique charm and style of Steven Seagal's movies\\
        A shy and introverted high school student \underline{struggling with} the effects of abuse at home & A mother of two \underline{seeking advice} on holistic approaches to boost her family's immune system\\
        a \underline{competitive} business strategist who works for a \underline{rival} smartphone company. & I am an emerging fashion designer who was deeply \underline{inspired by} Francine "Franc'" Pairon's legacy in the fashion industry and education.\\
        A social media influencer \underline{pushing against} the change, \underline{fearing} a loss of income prospects & A lucid dreamer who \underline{shares} their experiences and provides inspiration for new art projects\\
        A cattle rancher who \underline{struggles to} understand their daughter's choice and values & An artist who captures the unique beauty of snowboarding in their paintings and sculptures\\
        a retired school teacher who recently relocated to Mississippi & I am a budding botany enthusiast and amateur historian with a keen \underline{interest in} the flora of Africa and the work of early 20th-century botanists.\\
        A hardworking carpenter who \underline{competes} for the same construction projects & A medical doctor who is \underline{open to} alternative approaches and actively \underline{supports the collaboration} between the holistic nurse and music therapist\\
        A rival journalist who \underline{counters} debates with conservative views on penal justice & A frontline worker \underline{excited about} the potential improvements in their day-to-day tasks once the software deployment is complete\\
        a black female student studying law at Notre Dame. & A high school student \underline{interested in} history who is writing a report on Sweden for a school project\\
        A former cult member who experienced intense group pressure and conformity & A green investment specialist who can \underline{offer advice} on sustainable financing options and help the couple make responsible investment decisions\\
        A record label executive who sees the talent agent as a \underline{competitor} & A power plant manager \underline{seeking advice} on implementing carbon capture technology to reduce emissions\\
        A Bahian local who teaches history at a high school and is an \underline{activist against} racial inequality & A globetrotter who visits the village and promotes sustainable tourism, highlighting the cultural heritage of the site\\
        A \underline{competitive} small business owner in Singapore, not familiar with foreign exchanges & A holistic therapist who \underline{explores} natural remedies and lifestyle changes to alleviate psoriasis symptoms\\
         Another bathtub reglazing business owner in a nearby town, offering similar services and \underline{competing} for customers in the same area & As an expert in particle physics, particularly in theoretical and phenomenological analysis, \underline{my interest lies in} probing the properties of fundamental particles through high-energy collider experiments. The leap from previous colliders to the future electron-proton $(e^-p)$ collider facilities opens up a tantalizing field of research.\\
    \end{tabular}
    \caption{30 personas with worst (left column) and best (right column) alignment to the \textsc{Single-Label Dataset}. Annotations via Mistral, terms belonging to themes referenced in the text are underlined.}
    \label{tab:mistral_30}
\end{table*}

\begin{table*}
    \footnotesize
    \centering
    \begin{tabular}{p{7.5cm}|p{8cm}}
        A parent who is concerned about their child's digestive issues and \underline{seeks advice} on probiotics and gut-friendly foods & A traditional artist who \underline{questions} the authenticity and creativity of digital book cover designs\\
        A chef who \underline{shares insider tips} and techniques for cooking the dishes mentioned in the books & A resourceful scavenger who believes Carl is taking up valuable supplies\\
        A passionate gardener who \underline{shares tips and tricks} for finding the best gardening tools and ornaments & an ardent fan of stand-up comedy who appreciates Dusty Slay's unique humor and relatable stories.\\
        A photographer \underline{captivated by} the researcher's creative process and \underline{inspired to} capture their watercolor art in stunning photographs & I'm an aspiring novelist who focuses on themes of societal constructs and the impact of beauty standards on individuals, particularly women.\\
        A bed-and-breakfast owner who \underline{enjoys sharing} stories about the island with visitors & A pro-gun rights advocate who engages in heated debates with the son, \underline{providing contrasting perspectives} on the issue\\
        A grandparent who grew up baking bread from scratch and values the simplicity of traditional recipes & A sociologist studying the connection between cultural diversity and political alliances\\
        a long-term resident in Tampico, IL who loves DIY home improvement projects & A sociology student from Brazil, analyzing and discussing cultural differences and norms in their free time\\
        A 5-year-old boy with developmental coordination disorder who is eager to play sports with his friends & A reformed criminal who had a history of mental illness and can provide personal insight into the challenges faced by the defense\\
        A parent who is \underline{thankful for} the journalist's ability to provide accurate and easy-to-understand medical information for their children's health & A theology student studying the intersections between the Book of Mormon and other religious texts\\
        A proud homeowner who wants their property to be showcased in the best possible way & A ZANU-PF loyalist and ardent supporter of Robert Mugabe.\\
        A father of two, new to the neighborhood, who \underline{appreciates} the kindness and wisdom of older residents & A laissez-faire community moderator who values open dialogue and minimum intervention\\
         A \underline{curious} classmate who admires her knowledge about e-readers and often seeks her advice on choosing the right device & A retired history professor from Albi, France who has lived there for nearly 40 years and \underline{is slightly critical about} the recent surge of tourism in his beloved city.\\
        an Italian jazz lover who has been following Enrico Rava's music since the '70s & A local Cayman Islands resident \underline{with a disapproval towards} certain aspects of the local government's actions.\\
        A young child who loves Disney characters and wants a personalized artwork for their bedroom & A defense attorney who \underline{argues that} criminal justice reform measures \underline{could jeopardize} the rights of victims and hinder fair trials\\
        a world-traveling doctor who's \underline{searching for} her next adventure & A modern journalist who \underline{argues that} the media's role during the Civil War was \underline{exaggerated}\\
        A \underline{curious} traveler who wants to immerse themselves in the local culture and experience authentic Thai cuisine & A military general who believes in the deterring power of nuclear weapons\\
        An elderly patient with a long history of heart disease, determined to \underline{educate others} about the importance of heart health & An art major who believes that AI-generated art \underline{lacks} soul and creativity\\
        An accomplished travel blogger focusing on the hidden gems of San Francisco & A sociologist who argues that religion is not necessary for ethical behavior\\
        A fitness influencer promoting ergonomic practices for home office setups & A dedicated vegan and animal rights activist who's \underline{deeply against} any form of hunting.\\
        A homeowner considering installing energy-efficient appliances and \underline{seeking advice} on available incentives & A farmer from a neighboring country who is \underline{skeptical about} the benefits of free trade agreements\\
        A \underline{curious} and ambitious student eager to learn more about rocket propulsion and apply theoretical concepts in practical experiments & A Texas Longhorns fan who's always been \underline{skeptical of} Beard's coaching style\\
        A local historian and tour guide in Frascati & A European political analyst \underline{skeptical of} American gun policies\\
         A senior citizen who requires regular blood tests, \underline{appreciating} the gentle and skillful approach of the phlebotomist & A conservative, elderly man who \underline{is against} the legalization of marijuana due to his belief that all drugs can potentially harm society.\\
        A mother of two \underline{seeking advice} on holistic approaches to boost her family's immune system & A writer and blogger who frequently discusses Peter Dinklage's impact on disability representation in the entertainment industry\\
        A novelist who \underline{draws inspiration} from the serene beauty of gardens and often dedicates a chapter to it in their books & A \underline{disillusioned} economist who \underline{is willing to leak} confidential documents related to the economic crisis\\
        A farmer using telehealth services to \underline{consult with} veterinarians for livestock health management & A nuclear physicist who \underline{shares the concern} for the potential threats of nuclear weapons\\
        A wealthy businessman \underline{searching for} a luxurious mansion in a prestigious neighborhood & A computer science student who argues artificial intelligence proves determinism\\
        An entrepreneur who \underline{appreciates} the cook's creativity but also values preserving tradition & A political lobbyist who advocates for the continued use of landmines for national defense purposes\\
        An authoritative figure in the field of Geographic Information Systems (GIS) known for rigorous academic publications & A conspiracy theorist who \underline{challenges} the validity of forensic science and believes in alternative explanations\\
        A fellow sports mentor with a completely different training philosophy & A Sikh individual facing workplace discrimination due to their religious attire and practices\\
    \end{tabular}
    \caption{30 personas with worst (left column) and best (right column) alignment to the \textsc{Single-Label Dataset}. Annotations via Qwen, terms belonging to themes referenced in the text are underlined.}
    \label{tab:qwen_30}
\end{table*}

\clearpage

\subsection{Study 2 Results}

\begin{table}[t]
    \centering
    \begin{tabular}{c|r|r}
    Threshold & \# Clusters & \# Personas \\
    \hline \hline
    0.50 & 1,627 & 184,761 \\
    0.55 & 2,065 & 169,382 \\
    \textbf{0.60} & \textbf{2,180} & \textbf{138,519} \\
    0.65 & 1,676 & 87,653 \\
    0.70 & 702 & 30,954 \\
    0.75 & 102 & 3,613 \\
    \end{tabular}
    \caption{Similarity thresholds resulting in different clustering solutions in the persona-embedding space, together with the number of resulting clusters and the number of personas they include. Resulting cluster solution printed in bold.}
    \label{tab:cluster_solutions}
\end{table}

We set the similarity threshold for cluster formation to 0.6 - in combination with a minimum personas per cluster threshold of 25 - based on a comparison of the resulting clustering solutions using different similarity thresholds (Table \ref{tab:cluster_solutions}). We settle on this threshold value as it affords a high number of different clusters, which we think is necessary to account for the heterogeneity of persona descriptions, while still including a sufficiently high number of personas (70\% of all personas).

\begin{figure}[h!]
    \centering
    \includegraphics[width=\linewidth]{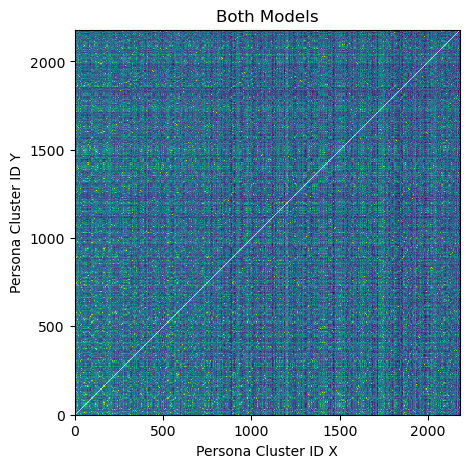}
    \caption{Íntra- and inter-cluster cosine distances of persona-space clusters measured in the persona embedding space shared by both models. Lighter-colored cells represent lower average distances between the respective clusters.}
    \label{fig:s3_pclusters_personaspace}
\end{figure}

Table \ref{tab:cluster_examples} provides examples for the three largest and three of the smallest clusters of our resulting cluster solution. As further confirmed in Figure \ref{fig:s3_pclusters_personaspace}, the resulting clusters are internally homogeneous (i.e., small average intra-cluster distance in the persona embedding space), as indicated by the light colors along the diagonal of Figure \ref{fig:s3_pclusters_personaspace}, as well as heterogeneous across different clusters (i.e., high(er) average inter-cluster distance in the persona embedding space), as indicated by the dark colors everywhere but on the diagonal of Figure \ref{fig:s3_pclusters_personaspace}.

\begin{figure}
    \centering
    \includegraphics[width=\linewidth]{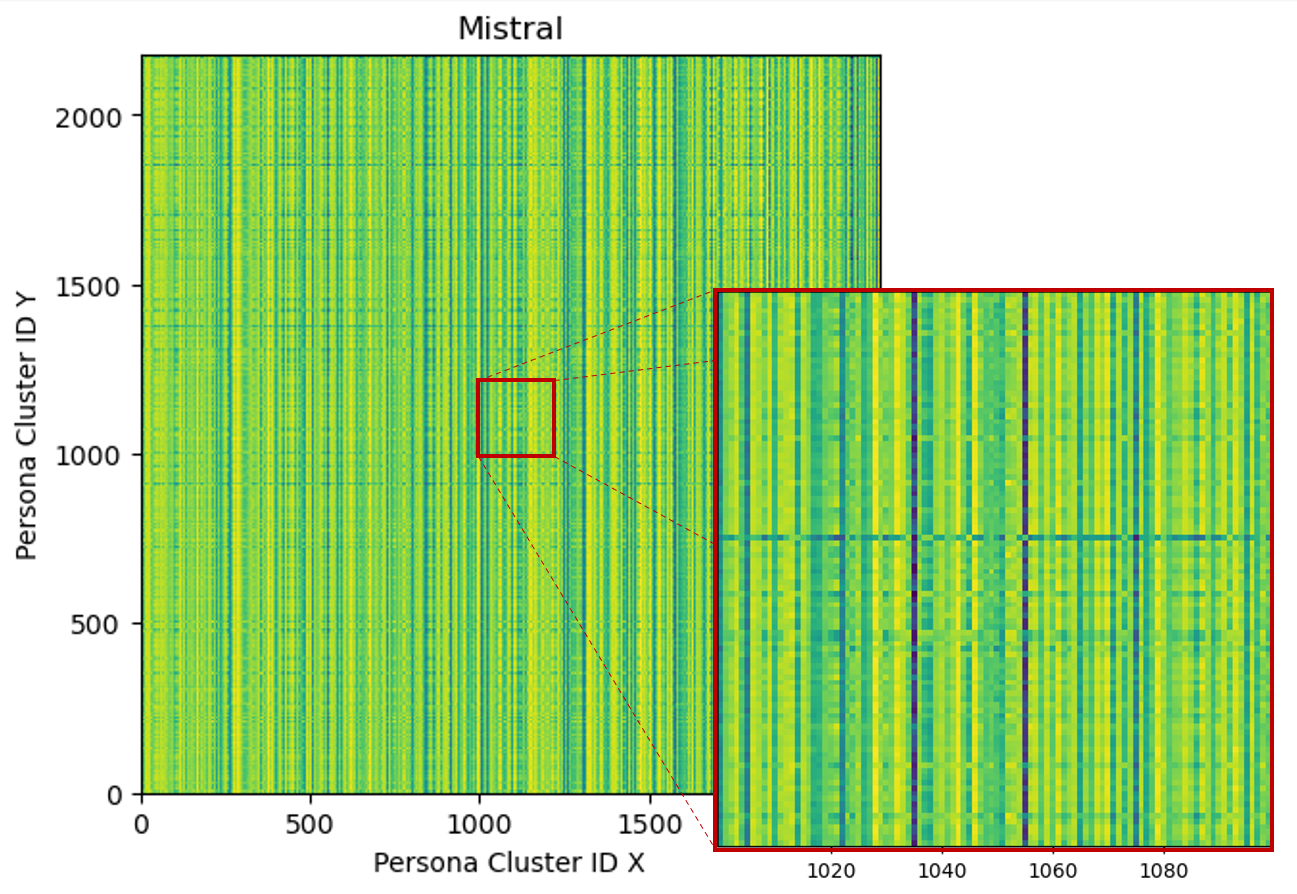}
    \caption{Intra- and inter-cluster cosine distances of persona-space clusters measured in label embedding space resulting from Mistral annotations. Values are normalized per row and lighter-colored cells represent lower average distances between the respective clusters. The inset zooms in on clusters with IDs from 1,000 to 1,100.}
    \label{fig:s2_pclusters_labelspace_mistral}
\end{figure}

Figure \ref{fig:s2_pclusters_labelspace_mistral} shows the lighter cell colors along the diagonal as a result of lower inter- than intra-cluster average distances for persona clusters in the label embedding space resulting from Mistral annotations.

\begin{figure}
    \centering
    \includegraphics[width=\linewidth]{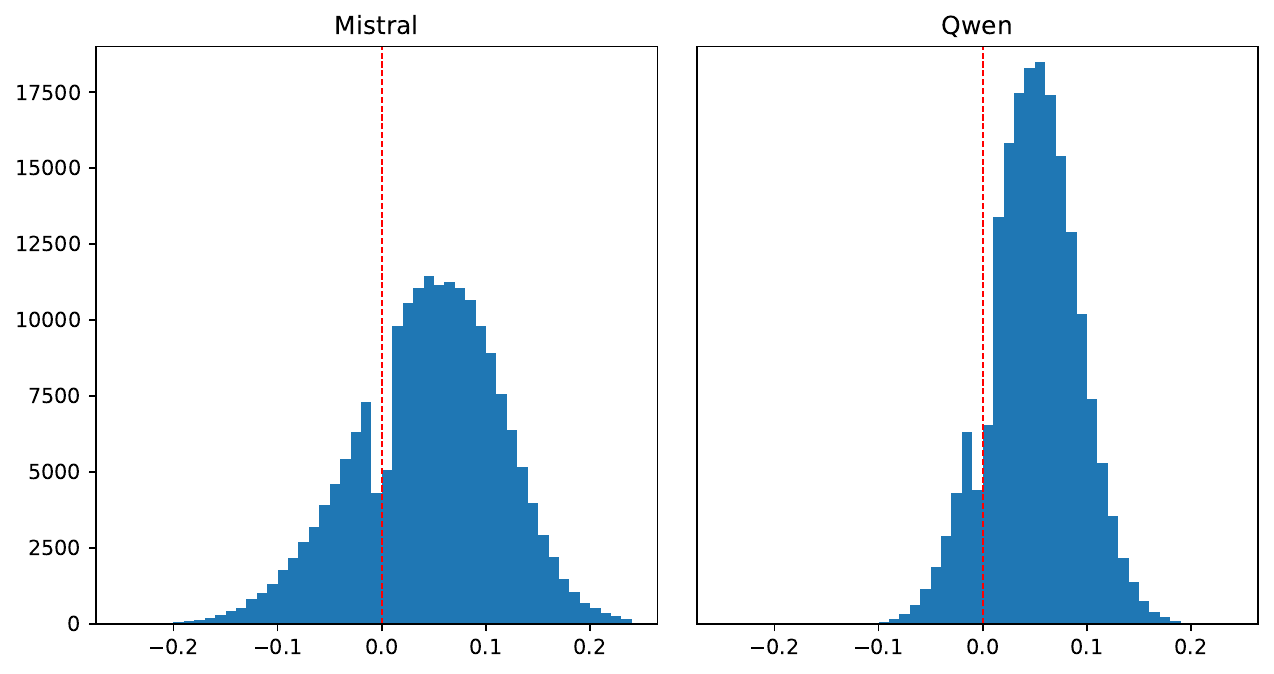}
    \caption{Histograms of Spearman correlation coefficients for pairwise distances measured in the persona and the label space. A single correlation coefficient represents the correlation between distances from a specific persona to every other persona in both spaces.}
    \label{fig:s2_hist_correlations}
\end{figure}

Figure \ref{fig:s2_hist_correlations} confirms the positive correlations between distances in the persona and the label space.

\clearpage

\begin{figure}
    \centering
    \includegraphics[width=\linewidth]{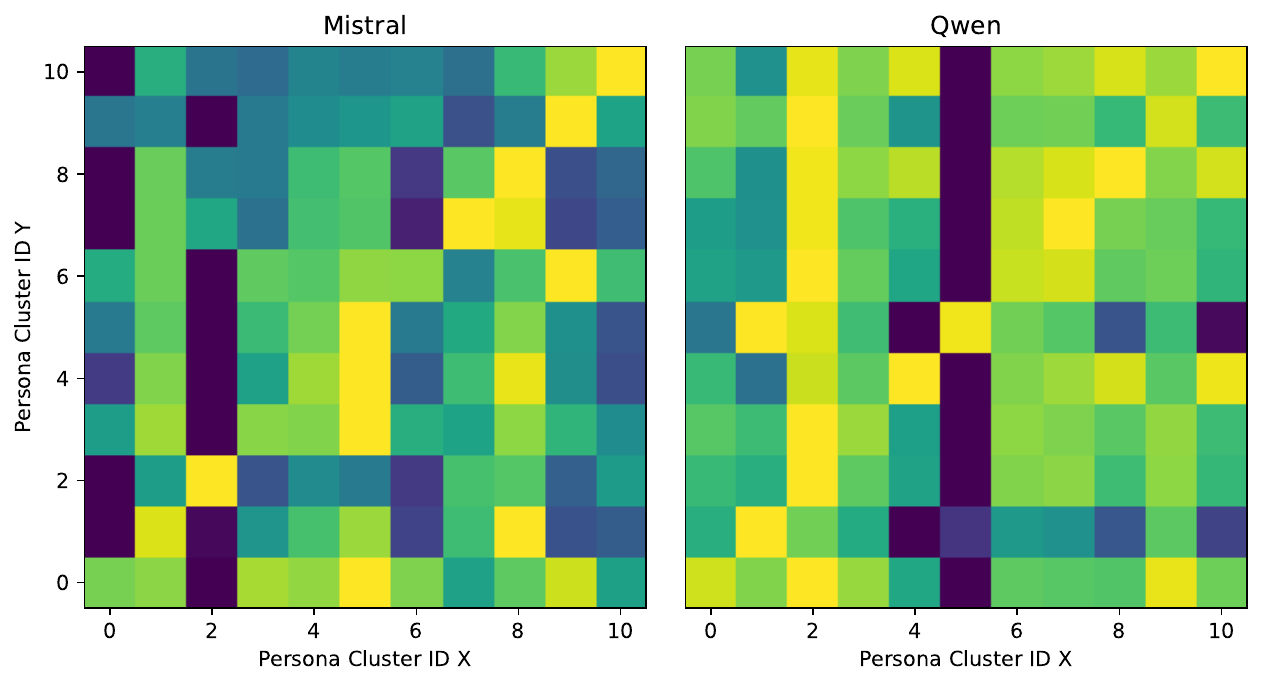}
    \caption{Intra- and inter-cluster cosine distances of label-space clusters measured in persona embedding space. Values are normalized per row and lighter-colored cells represent lower average distances between the respective clusters.}
    \label{fig:s2_lcluster_personaspace}
\end{figure}

Similar to the clustering done in the persona space, we also use k-means to cluster vectors in the label space, identifying clusters of personas that annotate alike and measuring their average intra- and inter-cluster distances in the embedding space built upon the persona descriptions, hoping to complement the insights derived from the persona-space clusters. Figure \ref{fig:s2_lcluster_personaspace} confirms the association between persona- and label space distances established through Figures \ref{fig:s2_pclusters_labelspace_qwen} and \ref{fig:s2_pclusters_labelspace_mistral}. For the eleven clusters found in the label embedding spaces of both models, we observe a tendency of cells along the diagonal to be brighter than the other cells in the same row, meaning that personas within the same label space cluster are more similar to each other than to personas in different label space clusters.

\begin{table*}
    \footnotesize
    \centering
    \begin{tabular}{p{1cm}|p{1cm}|p{4cm}|p{8cm}}
    ID & Size & Top Ten TF-IDF Cluster Terms & Three Random Cluster Personas \\
    \hline \hline
    0 & 1,393 & sports, athlete, player, basketball, professional, coach, tennis, athletes, sport, football & An athletics coach who focuses on talent development and has been tracking Alemitu’s career closely.; a freelance sportswriter; an esports fan who is confused about the appeal of physical sports.\\
    \hline
    1 & 1,349 & history, professor, historical, teacher, historian, literature, university, political, figures, specializing & A person who is fascinated by elaborate schemes and extraordinary behavior in history.; a literature professor at Lancaster University, United Kingdom.; A professor specializing in the specific subject area for which the content developer is creating material.\\
    \hline
    2 & 1,292 & journalist, political, news, reporter, politics, journalism, media, politician, commentator, reporting & a sports journalist for a local newspaper in Gloucester; A public relations specialist known for damage control in political controversies; An Orlando City SC player who values the journalist's support and uses their articles as motivation\\
    \hline
    ... & ... & ... & ... \\
    \hline
    2177 & 25 & fda, clinical, representative, trial, approving, drug, evaluating, responsible, reviewing, efficacy & A pharmaceutical industry regulator responsible for ensuring compliance with clinical trial protocols; A representative from the Food and Drug Administration (FDA) responsible for evaluating the safety and efficacy of the DNA sequencing technology; A representative from the regulatory agency responsible for overseeing and approving the use of CRISPR technology \\
    \hline
    2178 & 25 & friend, close, neighbor, old, johnson, colleague, dr, dale, long, longtime & an old friend and neighbor of Visanio Johnson.; a proud wife of a long-term friend of Dr. Robert Johnson; An old friend and golf buddy of Peter "Pete" Van Vooren, living in Sioux Falls, SD. \\
    \hline
    2179 & 25 & spiders, spider, fear, arachnophobic, arachnophobia, person, afraid, arachnophobe, arachnologist, bug & A homeowner who has a severe arachnophobia and wants to ensure a spider-free environment; An arachnophobic librarian; an arachnophobic tour guide in Australia \\
    \end{tabular}
    \caption{Cluster size, top ten TF-IDF terms and three randomly selected persona descriptions for the three largest and three of the smallest persona clusters.}
    \label{tab:cluster_examples}
\end{table*}

\begin{table*}
    \footnotesize
    \centering
    \begin{tabular}{p{4.9cm}p{1cm}|p{7.1cm}p{1cm}}
        AAE Instance & Diff. & Anti-Black Instance & Diff. \\
        \hline \hline
        All my \textit{n***} wear crowns, all my \textit{n***} bossed up, all my \textit{n***} got bad bitches don't none of us give a fuck. & -1.14 & Please save your black weaves for someone who gives a rat's ass & 1.78 \\
        \hline
        The way these \textit{n***} hate I hope they get a check for it & -1.13 & Is it even proven that we looked like big orangutans? They're most closely related to orangutans but the split happened an extremely long time ago, even before the split between gorillas and humans. Obviousy a large dark section of the population never made this split! & 0.73 \\
        \hline
        \textit{N***} rather be known and broke instead of low-key and getting chicken & -1.07 & Opps, Under our amazing President, Trump, gang members are gone. Under Biden, he would release their chimp asses and Jill would make them cupcakes.  awful ! God Bless President Trump & 0.53 \\
        \hline
    \end{tabular}
    \caption{AAE and anti-Black instances with largest difference in average toxicity ratings assigned by Black and conservative personas. Differences calculated as $\text{Diff}=\mu_{Q;B}-\mu_{Q;C}$ based on Qwen annotations. We replace all mentions of the n-word and its variants with \textit{n***}.}
    \label{tab:s2_m_diff_examples}
\end{table*}

\end{document}